\documentclass[10pt,twocolumn,letterpaper]{article}

\usepackage[pagenumbers]{cvpr} 

\usepackage{amsmath}
\usepackage{amssymb}
\usepackage{arydshln}
\usepackage{multirow}
\usepackage{bm}
\usepackage[dvipsnames]{xcolor}
\usepackage[ruled]{algorithm2e}
\usepackage{bbm}

\usepackage{float}
\restylefloat{table}

\bmdefine{\bMinus}{-}
\bmdefine{\bPlus}{+}

\def \bbR{{\mathbb{R}}}
\def \bb1{{\mathbbm{1}}}

\def \cL{{\mathcal{L}}}
\def \cN{{\mathcal{N}}}

\def \cX{{\mathcal{X}}}
\def \cY{{\mathcal{Y}}}

\def \bS{{\mathbf{S}}}

\def \bmp{{\bm{p}}}

\def \bmx{{\bm{x}}}
\def \bmy{{\bm{y}}}

\def \bu{{\boldsymbol\mu}}
\def \bS{{\boldsymbol\Sigma}}

\usepackage[dvipsnames]{xcolor}

\definecolor{cvprblue}{rgb}{0.21,0.49,0.74}
\usepackage[pagebackref,breaklinks,colorlinks,citecolor=cvprblue]{hyperref}

\def\paperID{*****} 
\def\confName{CVPR}
\def\confYear{2024}

\title{Domain Adaptation Using Pseudo Labels}

\author{Sachin Chhabra\\
Arizona State University\\
Tempe, Arizona, US\\
{\tt\small schhabr6@asu.edu}
\and
Hemanth Venkateswara\\
Georgia State University\\
Atlanta, GA, USA\\
{\tt\small hvenkateswara@gsu.edu}
\and
Baoxin Li\\
Arizona State University\\
Tempe, Arizona, US\\
{\tt\small baoxin.li@asu.edu}
}

\begin{document}
\maketitle
\begin{abstract}
In the absence of labeled target data, unsupervised domain adaptation approaches seek to align the marginal distributions of the source and target domains in order to train a classifier for the target. 
Unsupervised domain alignment procedures are category-agnostic and end up misaligning the categories. 
We address this problem by deploying a pretrained network to determine accurate labels for the target domain using a multi-stage pseudo-label refinement procedure. The filters are based on the confidence, distance (conformity), and consistency of the pseudo labels.
Our results on multiple datasets demonstrate the effectiveness of our simple procedure in comparison with complex state-of-the-art techniques.
\end{abstract}    
\section{Introduction}
\noindent 
Supervised learning with deep convolutional neural networks has achieved state-of-the-art results on many challenging computer vision tasks \cite{he2016deep,redmon2016you,ronneberger2015u,patchswap,patchrot}. 
However, training large neural networks requires massive amounts of labeled data like the ImageNet. 
Such datasets are limited to a few applications and are expensive to create for most real-world applications. 
Transfer learning techniques have been used to adapt deep neural networks trained for one application to other related applications that do not have large labeled datasets \cite{tan2018survey}. 
Domain adaptation is a special case of transfer learning where a network trained on a source domain (dataset) is adapted to the target domain (dataset) in the absence of labeled target data \cite{venkateswara2017deepSPM}. 
Domain adaptation is applied to variety of problems like classification \cite{ganin2016domain}, detection \cite{chen2018domain}, regression \cite{lei2021near}, pose estimation \cite{zhang2019unsupervised}. Some approaches focus on speeding the adaptation process \cite{li2021faster}. 
In this paper, we tackle the problem of adapting classifier to target domain.

Unsupervised domain adaptation models usually focus on aligning the data distributions of the source and target domains when training adaptive classifiers. 
When the domains are aligned, a classifier trained on the source can classify the unlabeled target data. 
Domain alignment in computer vision is found majorly in two forms, feature alignment \cite{ganin2016domain,long2015learning,chhabra2021glocal} and pixel-level alignment \cite{hoffman2018cycada,russo2018source}. 
Feature alignment leverages the power of a deep neural network to extract image features from the source and target. These models align the marginal distributions of the features before training a classifier. 
Pixel-level alignment takes advantage of the generative capabilities of deep networks to translate images from one domain to another before training a classifier. 
These alignment approaches popularly include adversarial methods \cite{ganin2016domain,tzeng2017adversarial}, cyclic generative models \cite{hoffman2018cycada,russo2018source} and distribution alignment metrics \cite{long2015learning,shen2018wasserstein}. 
Even with a slew of complex alignment approaches, the best domain alignment is arguably achieved by the deep network itself.
This is evident from the fact that the state-of-the-art approaches in domain adaptation use the latest deep neural networks to demonstrate their results. 

Category agnostic domain alignment (marginal distribution alignment) does not guarantee accurate target predictions with a source classifier \cite{long2017deep}. 
However, the problem of category-wise domain alignment appears to be intractable without the target data labels. 
A pretrained network fine-tuned with a labeled source dataset can be used to obtain reliable pseudo labels for the target. 
With stringent criteria for selecting the most reliable pseudo labels, we propose to gradually coerce the source classiﬁer to accurately classify the target.

\begin{figure*}[t]
    \begin{center}
       \includegraphics[width = \textwidth]{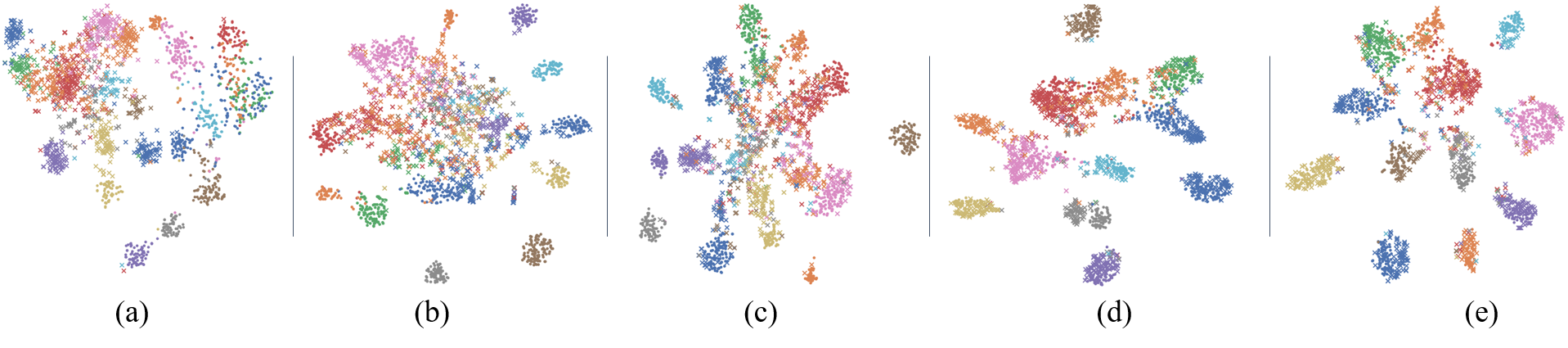}
    \end{center}
 \caption{Visualization of image features using t-SNE. The source domain is represented using $\bullet$ and the target domain using $\mathbf{\times}$. Different colors depict different categories. Best viewed in color. (a) ResNet101 trained on ImageNet (b) Pretrained ResNet101 over-fitted with the source domain  (c) DANN \cite{ganin2016domain} (d) Trained on source and target domain (e) Our method.}
\label{Fig:tsne}
\end{figure*}

In Figure \ref{Fig:tsne} we illustrate the promise in our method with a preliminary result over the VisDA \cite{peng2017visda} dataset. 
A pretrained deep network like ResNet101 extracts generic features that clusters into categories as in Figure \ref{Fig:tsne}(a). 
When ResNet101 is re-trained with only the source data, the network overfits and only clusters the source data: Figure \ref{Fig:tsne}(b). 
A marginal domain alignment like the DANN \cite{ganin2016domain} provides only marginal improvement in the clustering: Figure \ref{Fig:tsne}(c). 
Optimal clustering for the source and target is obtained when the network is trained with labeled data for both the source and target: Figure \ref{Fig:tsne} (d). 
A similar clustering can also be obtained by gradually incorporating reliable pseudo labels into the labeled data as with our method: Figure \ref{Fig:tsne}(e). 

We demonstrate how a simple pseudo-label filtering procedure yields a competitive baseline for domain adaptation. Our proposed approach is simple and efficient and achieves comparable performance.
It is counter-intuitive how this throwback principle of multi-stage refinement of pseudo labels yields astounding results. 
The primary challenge with using pseudo labels is the issue of confirmation bias that can arise with noisy labels \cite{tarvainen2017mean}. 
We develop a rigorous 3-stage pseudo label refinement accounting for confidence, conformation, and consistency. 
In the first stage, we select pseudo labels the model has \emph{Confidence} about. 
In the second stage, the \emph{Conformity} filter selects only non-spurious samples that belong to the distribution. 
The third stage ensures \emph{Consistency} in the pseudo label across epochs. 
Our filtering criteria are based on (i) output confidence, (ii) distance from mean of a Gaussian distribution and (iii) consistency. 
We proposed to use these filters in conjunction for unsupervised domain adaptation. 

\section{Related Work} 
\noindent Unsupervised domain adaptation approaches aim to align the source and target distributions. 
One of the most popular methods to reduce domain disparity is adversarial feature alignment  \cite{ganin2016domain,tzeng2015simultaneous}, where an adversarial discriminator is used to align the features from the two domains using the principle of a GAN \cite{goodfellow2014generative}. 
Adversarial alignment has been refined with multiple feature extractors and classifiers in ADDA \cite{tzeng2017adversarial}, MADA \cite{Pei2018multi} and MCD \cite{saito2018maximum}. 
Adversarial alignment with category and domain Mixup \cite{zhang2018mixup} was used in \cite{wu2020dual}. 
Adversarial attacks were utilized to improve model's generalization performance in \cite{li2021divergence}. Adversarial dropout was used to generate domain-invariant features in  \cite{lee2019drop}, and TADA transfers attention patterns from source to the target \cite{wang2019transferable}. 
These approaches attempt to globally align the image features with little control over the category-level alignment. 

Training using pseudo labels is popular in semi-supervised learning where confident pseudo labels from the unlabelled data are added to the labeled set \cite{lee2013pseudo}. 
Pseudo-labels have also been successfully adopted for unsupervised domain adaptation. 
Co-training procedures have explored gradually augmenting the labeled set with pseudo-labeled target samples \cite{chen2011co}. 
MSTN \cite{xie2018learning} learns semantic representations for unlabeled target samples by aligning the centroids of the source and those obtained by pseudo target labels. 
Target samples that retain their pseudo labels across epochs are used for training in \cite{zhang2018collaborative}. 
A procedure to align the category clusters of source and target samples has been developed using pseudo labels in \cite{chen2019progressive}. 
Adaptation of source and target domains at output-level was proposed in \cite{chhabra2023generative}.

These techniques rely on the pseudo labels which is not always accurate. 
Pseudo label accuracy is the key to a good target classifier because noisy pseudo labels can mislead the model and result in confirmation bias \cite{tarvainen2017mean}. 
A procedure for authenticating pseudo labels has been explored in \cite{saito2017asymmetric} using two networks. 
The confident pseudo labels are used to train a third network for domain adaptation. 
In \cite{chen2020adversarial}, the classifier and the adversarial discriminator have been combined to generate a confusion matrix that can estimate the correctness of the pseudo label. 
A conditional GAN and a classifier are used in tandem to refine pseudo labels in \cite{morerio2020generative}. 

Our pseudo-label-based approach is orthogonal to the methods discussed so far. 
Rather than correcting the pseudo labels, we use a simpler procedure of filtering to select the most accurate pseudo labels. 
Selecting pseudo labels is generally based on the confidence criteria (\cite{chen2011co,xie2018learning,zou2018unsupervised}). 
We introduce more stringent criteria to arrive at a more reliable set of pseudo labeled samples. 
Finally, we use the selected pseudo labels to gradually transform the source classifier.  

\section{Approach}
\noindent In this section we outline the problem statement, develop a pseudo label generator and present our multi-stage pseudo label filtering strategy. 

\subsection{Problem Statement}
\noindent Let $D_s = \{(\bmx_i^s, \bmy_i^s)\}_{i=1}^{n_s}$ be the source domain consisting of $n_s$ labeled images where $\bmx^s \in \cX^s$ and $\cX^s$ is the space of source images, $\bmy^s\in\cY=\{0,1\}^C$ are one-hot representations of the image labels in binary label space $\cY$. There are $C$ classes and the corresponding component in $\bmy = [y^1,\ldots,y^C]^\top$ is 1, the rest are 0s. 
Likewise the target domain is denoted as $D_t = \{(\bmx_i^t)\}_{i=1}^{n_t}$ consisting of $n_t$ unlabeled samples with $\bmx^t \in \cX^t$. The target labels $\bmy^t\in\cY^t$ are not available but it is known that they belong to the same label space as the source, $\cY$.  It is assumed that the source dataset is sampled from the joint distribution $p_s(\bmx, \bmy)$ and the target samples are sampled using distribution $p_t(\bmx)$. The goal of unsupervied domain adaptation is to train a classifier model using $D_s$ and $D_t$ to predict the labels of the target dataset labels $\{\hat{\bmy}^t_i\}_{i=1}^{n_t}$. Although the source and target datasets have the same label space, a classifier trained using the source dataset will perform poorly at predicting the target labels due to the distribution difference between the source and target; $p_s \neq p_t$. 

\begin{figure*}[t]
    \begin{center}
       \includegraphics[width = 0.8\textwidth]{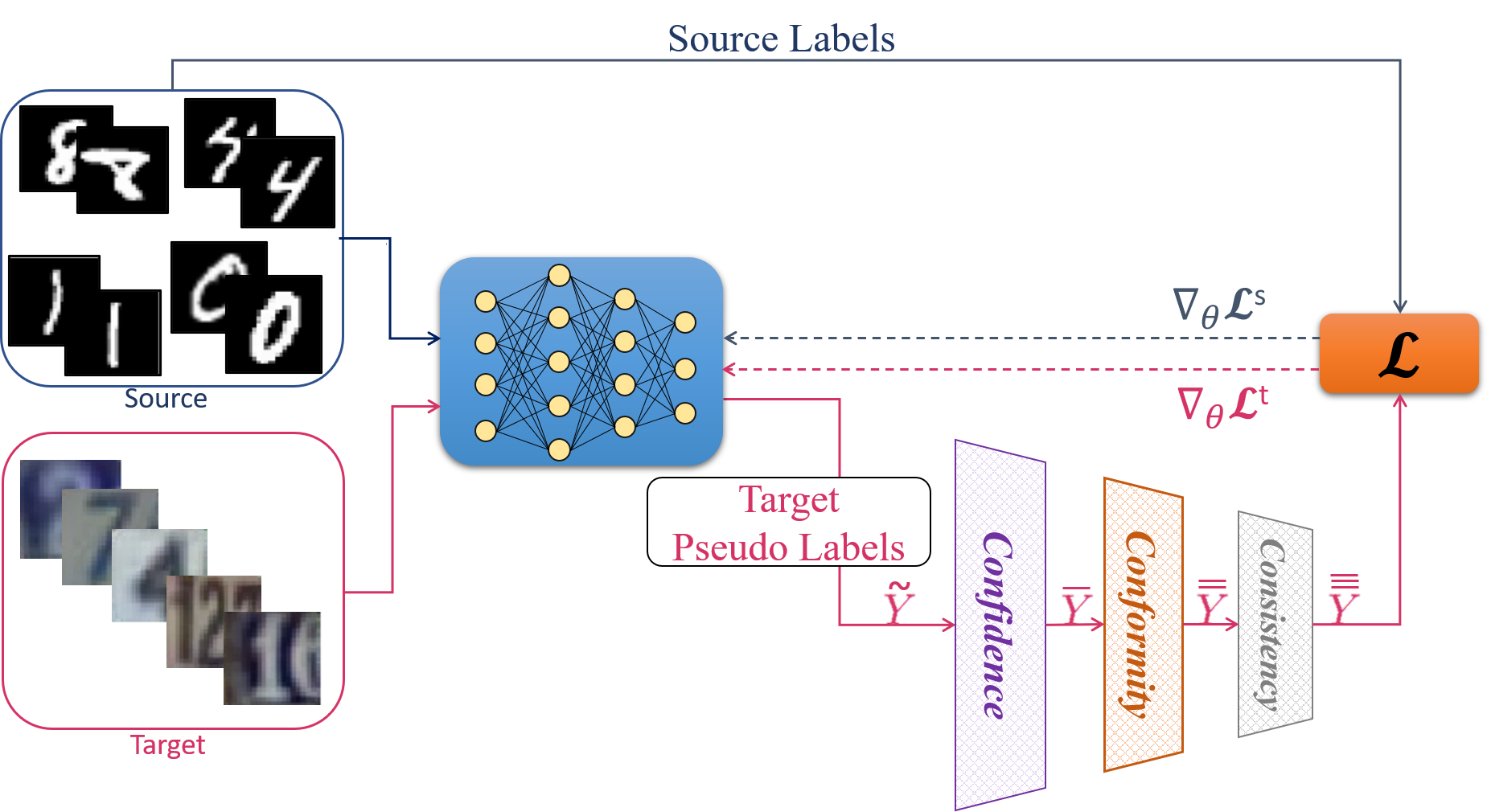}
    \end{center}
\caption{Our proposed framework. Source and target samples are input to the network. 
The network is trained on source data in supervised way. For target adaptation, we train the network on the target pseudo labels. The obtained pseudo labels for the target are filtered to get the the most precise subset using \emph{Confidence}, \emph{Conformity} and \emph{Consistency} filters before they are promoted to the labeled dataset to be used for supervised training. \emph{Confidence} filter validates the sample based on their output confidences. \emph{Conformity} filter verifies that the sample lies within the Gaussian region. \emph{Consistency} filter ensures that the sample has been consistent with its output.
Best viewed in color.
}
\label{model}
\end{figure*}

\subsection{Pseudo Label Generator} 
\noindent Standard unsupervised domain adaptation algorithms train a classifier using labeled source data and align the marginal distributions of the source and target datasets \cite{ganin2016domain}. 
We rely upon the feature extraction capabilities of the pre-trained network to perform the domain alignment.  
However, we need to provide supervision in order to coerce a pre-trained network to accurately classify unlabeled target data. 
We propose to qualify and promote select target samples and their assigned pseudo labels to the supervised dataset to be used for training the network. 
Our framework to evaluate the goodness of the pseudo labels is based the following criteria: (i) strength of the pseudo label, or a measure of confidence in the pseudo label, (ii) likelihood that the target sample is in-distribution (as opposed to out-of-distribution), (iii) accounting for noise in label assignment. 
We will elaborate upon these criteria in Section \ref{Sec:PsLF}. 

We use a cluster framework to evaluate the goodness of the pseudo labels because of its intuitiveness although any generative classifier framework can be used. 
Gaussian Mixture based classifiers provide a robust probabilistic generative framework for pseudo label evaluation and have demonstrated success in deep learning applications \cite{wan2018rethinking}. 

We design a deep neural network based feature extractor $G_\theta$ with parameters $\theta$, that outputs features $G_\theta(\bmx)$ given an image $\bmx$. 
A Gaussian Mixture loss is imposed on the image features $G_\theta(\bmx)$ and the deep neural network is trained to cluster the features with their category labels. 
While training the network on target data, it bring the target samples close to the source class centroids and leads to domain alignment without any explicit domain alignment loss.  
During training, the model updates the centroid $\bu$ and covariance matrix $\bS$ for each class. The objective function for training a standard Gaussian Mixture model with labeled samples $\{\bmx_i, \bmy_i\}_{i=1}^n$ is,


\begin{footnotesize}
\begin{flalign} 
\cL_{gm} =& -\frac{1}{n}\sum_{i=1}^n \bb1(y_i^k=1)\text{log}\frac{\cN\big(G_\theta(\bmx_i);\bu_{k}, \bS_{k}\big)p(k)}{\sum_{c=1}^C\cN\big(G_\theta(\bmx_i);\bu_{c}, \bS_{c}\big)p(c)} \notag\\
&-\gamma\sum_{i=1}^n\bb1(y_i^k=1)\text{log}\cN\big(G_\theta(\bmx_i);\bu_{k}, \bS_{k}\big)p(k), \label{Eq:GMLoss}
\end{flalign}
\end{footnotesize}
\noindent where the first term is the posterior probability that $G_\theta(\bmx_i)$ belongs to cluster $k$ and the second term is the likelihood that $G_\theta(\bmx)$ belongs to the Gaussian Mixture and $\gamma$ controls the importance of the likelihood term.  
$\bb1(.)$ is an indicator function. 
$\cN\big(G_\theta(\bmx_i);\bu_{c}, \bS_{c}\big) \propto \text{exp}(-d_c^i)$ is the Gaussian probability where $\bS_c$ is the covariance matrix of class $c$ and $\bu_c$ is the mean of the deep features/centroid for class $c$. 
$d_c^i$ is the Mahalanobis distance given by $\frac{1}{2}(G_\theta(\bmx_i) - \bu_c)^\top\bS_c^{-1}(G_\theta(\bmx_i) - \bu_c)$.  
Following \cite{wan2018rethinking}, we simplify above equation by restricting $\bS_c$ to diagonal matrices and set the prior probabilities to constants $p(c) = \frac{1}{C}$. 
We estimate $\bu_c$, $\bS_c$ and $\theta$ through backpropogation. 
With a trained Gaussian Mixture network we obtain the prediction probabilities of a target sample using,
\begin{flalign} 
p(y_i^k=1|G_\theta(\bmx_i)) = \frac{|\bS_{k}|^{-\frac{1}{2}}\text{exp}^{-d_k^i}}{\sum_c|\bS_c|^{-\frac{1}{2}}\text{exp}^{-d_c^i}}.
\label{Eq:GMM4}
\end{flalign}

\subsection{Target Supervision Schedule} 
\label{Sec:TSschedule}
\noindent According to Eq. \ref{Eq:GMM4}, every target sample $\bmx$ is assigned a probability vector $\bmp = [p^1, p^2, \ldots, p^C]^\top$, where $p^c$ is the probability that $\bmx$ belongs to category $c$. 
The data point $\bmx$ is assigned a pseudo label based on the maximum probability category. 
We propose to augment the set of labeled data with an increasing number of pseudo-labeled target samples and retrain the classifier iteratively. 
We hypothesize that this procedure will gradually modify the classifier to improve target classification accuracy. 
At the beginning of the training process, most of the pseudo labels are likely inaccurate because the classifier has been trained using only the source data. 
It is apparent that we need to retain only the most accurately-assigned pseudo labels before adding them to the labeled set. 

Beginning with zero target samples in the first iteration of training, we intend to add all the target data and their pseudo labels to the pool of labeled data by the end of the training. 
We propose a category-wise scheduler that gradually increases the fraction of pseudo-labeled samples that are promoted to the training stage for each category. 
A single selection criterion for filtering the target data may lead to a class imbalance with compact clusters in the Gaussian Mixture yielding more samples. 
For a total of $T$ epochs, at every epoch $t$ we sort the target samples in a descending order based on the maximum probability category and select the top $\lfloor\frac{n_t}{C}.s(t)\rfloor$ samples from every category. 
We generate the target pseudo label set $\tilde{Y} = \{\tilde{\bmy}_i\}_{i=1}^{\tilde{n}}$ where, $\tilde{\bmy}$ is a one-hot vector pseudo label and $s(t)$ is the fraction of samples selected, where, 
\begin{flalign} 
s(t)=\frac{1+\cos(\pi-\frac{\pi t}{T})}{2}.
\label{cosr}
\end{flalign}
$s(.)$ is a nonlinear monotonic increasing function that ranges between $[0,1]$ as the epoch number $t$ ranges from $\{1,2,\ldots,T\}$ and $\tilde{n}$ is the total number of pseudo labels selected. 
Small values for $s(t)$ in the initial stages (zero slope) helps generate reliable pseudo labels compared to a linear schedule with $s(t) = \frac{t}{T}$. 

\begin{table*}[t]
\footnotesize
\begin{center}
 \begin{tabular}{l|c c c c c c c c c c c c | c} 
 \toprule
 Method & Plane & Bcycl & Bus & Car & Horse & Knife & Mcycl & Person & Plant & Sktbrd & Train & Truck & Mean\\ 
 \midrule
Source only & 55.1 & 53.3 & 61.9 & 59.1 & 80.6 & 17.9 & 79.7 & 31.2 & 81.0 & 26.5 & 73.5 & 8.5 & 52.4\\
DANN\cite{ganin2016domain} & 81.9 & 77.7 & 82.8 & 44.3 & 81.2 & 29.5 & 65.1 & 28.6 & 51.9 & 54.6 & 82.8 & 7.8 & 57.4 \\
DAN\cite{long2015learning} & 87.1 & 63.0 & 76.5 & 42.0 & 90.3 & 42.9 & 85.9 & 53.1 & 49.7 & 36.3 & 85.8 & 20.7 & 61.1 \\
MCD\cite{saito2018maximum} & 87.0 & 60.9 & 83.7 & 64.0 & 88.9 & 79.6 & 84.7 & 76.9 & 88.6 & 40.3 & 83.0 & 25.8 & 71.9\\ 
ADR\cite{saito2018adversarial} & 87.8 & 79.5 & 83.7 & 65.3 & 92.3 & 61.8 & 88.9 & 73.2 & 87.8 & 60.0 & 85.5 & 32.3 & 74.8\\ 
DMRL\cite{wu2020dual} &  - &-  &  -&-  &  -&  -& - & - & - & - & - & - &  75.5\\
CDAN+BSP\cite{chen2019transferability} & 92.4 & 61.0  & 81.0  & 57.5 & 89.0  & 80.6 & 90.1 & 77.0  & 84.2 & 77.9 & 82.1 & 38.4 & 75.9\\
SWD\cite{lee2019sliced} & 90.8 & 82.5 & 81.7 & 70.5 & 91.7 & 69.5 & 86.3 & 77.5 & 87.4 & 63.6 & 85.6 & 29.2 & 76.4\\ 
ALDA\cite{chen2020adversarial} & \textbf{93.8} & 74.1 & 82.4 & 69.4 & 90.6 & 87.2 & 89.0 & 67.6 & 93.4 & 76.1 & 87.7 & 22.2 & 77.8\\
PANDA\cite{hu2020panda} & 90.9& 50.5 &72.3 &82.7 &88.3& \textbf{88.3}& 90.3 &79.8& 89.7& 79.2& 88.1 &39.4& 78.3\\
CRST+R$_{EBM}$\cite{liu2021energy} & 90.3 &  82.6 &  72.4 &  71.7 &  87.6 &  81.8 &  85.4 &  80.8 &  87.1 &  \textbf{89.9} &  83.6 &  71.5 &  80.2\\
TPN\cite{pan2019transferrable} & 93.7 & 85.1 & 69.2 & 81.6 & \textbf{93.5} & 61.9 & 89.3 & 81.4 & 93.5 & 81.6 & 84.5 & 49.9 & 80.4\\
DTA\cite{lee2019drop} & 93.7 & 82.2 & \textbf{85.6} & \textbf{83.8} & 93.0 & 81.0 & \textbf{90.7} & \textbf{82.1} & \textbf{95.1} & 78.1 & 86.4 & 32.1 & 81.5\\
CLS\cite{liu2021adversarial} & 92.6 & 84.5 & 73.7 & 72.7 &88.5 & 83.3 & 89.1 & 77.6 &89.5 & 89.2 & 85.8 & \textbf{72.0} & 81.6\\
DAPL(Ours)  & 93.4 & \textbf{89.6} & 84.7 & 81.5 & \textbf{93.5} & 82.8 & 90.6 & 78.2 & 92.5 & 64.4 & \textbf{91.9} & 39.9 & \textbf{81.9}\\
\bottomrule
\end{tabular}
\end{center}
\caption{Classification accuracies of different approaches on the VisDA dataset (ResNet-101).}
\label{visdaresults}
\end{table*}

\subsection{Pseudo Label Filtering} 
\label{Sec:PsLF}
\noindent In this subsection we outline how we evaluate the goodness of a pseudo label before we promote the target sample to the labeled dataset. \\ 

\noindent \textbf{\emph{Confidence}}: 
Using the \emph{Confidence} criterion we propose to retain only those target samples where the classifier has the highest confidence in the prediction. 
To do so, we select samples whose maximum predicted probability in $\bmp$ is above a cutoff threshold $p_{\tau}$. 
Since the classifier is designed to improve with training, we execute scheduled filtering where we increase the cutoff threshold monotonically from $\tau_s$ to $\tau_f$. 
For a total of $T$ epochs, at every epoch $t$, the cutoff probability threshold is given by $p_{\tau}(t) = \tau_s + (\tau_f - \tau_s).s(t)$, where $s(t)$ is from Eq. \ref{cosr}. $\tau_f=0.99$ and $0\leq\tau_s\leq\tau_f$ are constants to ensure $p_{\tau} \in [0,1]$. 
$\tau_s$ is adjusted to ensure there are sufficient samples for subsequent filtering. 
The output of the confidence filtering is the set of target pseudo labels $\bar{Y} = \emph{Confidence}(\tilde{Y})$ where $\tilde{Y}$ is obtained from the Target Supervision Schedule discussed in the subsection \ref{Sec:TSschedule}. 
We represent different levels of pseudo label filtering with different accents over $Y$.\\

\noindent \textbf{\emph{Conformity}}: 
It is possible to have high confidence target samples that do not conform to a compact Gaussian cluster model. 
For example, outlier target samples that are not close to anyone cluster center could have high confidence in prediction when they are relatively close to one cluster center compared to other cluster centers. 
The likelihood loss in Eq. \ref{Eq:GMLoss} (2nd term), reduces the chance of such outliers by enforcing compact clusters. 
Using \emph{Conformity} we further filter the pseudo labels to lie close to cluster centers in all the dimensions. 
In other words, we propose to reject samples that do not conform to a compact Gaussian Mixture model. 
To identify the non-conforming samples we determine the matrix $Z \in \bbR^{C\times C}$ where, $z_{i,j} \in Z$ is defined as, 
\begin{flalign}
z_{ij} = \left|\frac{\bu^k_i - \bu^k_j}{\sigma_i^k}\right|,
\end{flalign}
the maximum component-wise $z$-distance from cluster $i$ to cluster $j$ across all the $d$-components. 
Here, $\bu^k$ is the $k$-th component of the $d$-dimension vector $\bu$ and $\sigma_i^k$ is the standard deviation of the $k$-th component of cluster $i$. 
We then define the threshold $z_{th} = \min\{z_{i,j}\} ~\forall ~i,j$. 
In order to filter out the outliers, we identify $z$-distance values for all target samples and test them against the $z_{th}$. We define the $z$-distance $z_{\bar{\bmx}}$ of a target sample $\bar{\bmx}$ as the maximum component-wise distance across the $d$ dimensions from its closest cluster $c$,
\begin{flalign}
z_{\bar{\bmx}} = \max \left|\frac{G_\theta(\bar{\bmx}) - \bu_c}{\sigma_c}\right|
\end{flalign}
Given a target sample $\bar{\bmx}$, if any of its $z$-distance is greater than $z_{th}$, it implies that $\bar{\bmx}$ is not close enough to its assigned cluster center and perhaps closer to another cluster in some other dimension. 
We reject such samples. 
The output of the second level of filtering is the set of target pseudo labels $\bar{\bar{Y}} = \emph{Conformity}(\bar{Y})$. \\

\noindent \textbf{\emph{Consistency}}: In this filtering, we reject target samples that are not consistent with their label predictions over time. 
As the parameters of the feature extractor are updated, the Gaussian Mixture takes shape with samples realigning themselves in the feature space. 
It is likely that target samples may accidentally clear the confidence and conformity filters and yet be incorrectly labeled \cite{zhoutime}. 
Hence, time-consistency is an important factor. 
We wait for one epoch before we can promote a target sample to the labeled set to be used for training in backpropagation. 
A target sample clears this filter if its pseudo label is consistent across 2 successive epochs and it has cleared the first two filters as well.
The output of the third level of filtering is the set of target pseudo labels $\bar{\bar{\bar{Y}}} = \emph{Consistency}(\bar{\bar{Y}})$. The three filters can be applied in any order and together they achieve a refined set of pseudo labels for the target samples. 
We illustrate the effect of filtering in supplementary material using tSNE plots of target samples. 



\begin{table*}[t]
\footnotesize

\begin{center}

 \begin{tabular}{l|c c c : c c c : c c c : c c c | c}
 \toprule
 Source & \multicolumn{3}{c:}{Ar} & \multicolumn{3}{c:}{Cl} & \multicolumn{3}{c:}{Pr} & \multicolumn{3}{c|}{Rw} & \multirow{2}{*}{Mean} \\ 
 \cline{2-13}
  Target & Cl & Pr & Rw & Ar & Pr & Rw & Ar & Cl & Rw & Ar & Cl & Pr\\ 
  
 \midrule
Source only & 34.9&	50.0 &	58.0 &	37.4&	41.95&	46.2&	38.5&	31.2&	60.4&	53.9&	 41.2&	59.9&	46.1\\
DAN\cite{long2015learning} & 43.6&	57.0 &	67.9&	45.8&	56.5&	60.4&	44.0 &	43.6&	67.7&	63.1&	51.5&	74.3&	56.3\\
DANN\cite{ganin2016domain}&45.6&	59.3&	70.1&	47.0 &	58.5&	60.9&	46.1&	43.7&	68.5&	63.2&	51.8&	76.8&	57.6\\
CDAN+E\cite{long2018conditional}&50.7&	70.6&	76.0 &	57.6&	70.0 &	70.0 &	57.4&	50.9&	77.3&	70.9&	56.7&	81.6&	65.8\\
CDAN+BSP\cite{chen2019transferability}&52.0 	&68.6	&76.1	&58.0 	&70.3	&70.2	&58.6	&50.2	&77.6	&72.2	&59.3	&81.9	&66.3\\
ALDA\cite{chen2020adversarial}& \textbf{53.7}	&70.1	&76.4	&60.2	&72.6	&71.5	&56.8	&51.9	&77.1	&70.2	&56.3	&82.1	&66.6\\
SymNets\cite{zhang2019domain}&47.7&	72.9&	78.5&	\textbf{64.2}&	71.3&	74.2&	64.2&	48.8&	79.5&	\textbf{74.5}&	52.6&	82.7&	67.6\\
TADA\cite{wang2019transferable}&53.1	&72.3	&77.2	&59.1	&71.2	&72.1	&59.7	&\textbf{53.1}	&78.4	&72.4	&\textbf{60.0}	&82.9	&67.6\\
DAPL(Ours)  & 53.2 & \textbf{78.0} & \textbf{80.0} & 63.7 & \textbf{75.6} & \textbf{76.4} & \textbf{65.3} & 52.3 & \textbf{81.9} & 71.6 & 56.6 & \textbf{83.6} & \textbf{69.9}\\
\bottomrule
\end{tabular}
\end{center}
\caption{Classification accuracies of different approaches on the Office-Home dataset (ResNet-50).}
\label{oh_results}
\end{table*}

\subsection{Supervised Training} 
\noindent The multi-stage pseudo label refinement yields $n'$ target samples and their corresponding pseudo labels $\{(\bar{\bar{\bar{\bmx}}}_j^t, \bar{\bar{\bar{\bmy}}}_j^t)\}_{j=1}^{n'}$, where $n'<n_t$. 
The labeled source dataset is augmented with the pseudo label dataset and used in training the Gaussian Mixture model by minimizing Eq. \ref{Eq:GMLoss} using backpropagation. 
The objective term considering a sample data point from each of the domains is, 
\begin{flalign}
   \cL = \cL_{gm}(\bmx^s; \bmy^s; \theta) + \lambda  \cL_{gm}( \bar{\bar{\bar{\bmx}}}^t; \bar{\bar{\bar{\bmy}}}^t; \theta).
\label{Eq:ent_min}
\end{flalign}

Since the selected target examples are the most confident, conforming, and consistent, the loss from these samples is nearly zero. 
In turn, the gradient of the loss w.r.t. these samples is also nearly zero.

Therefore the network does not update itself with information from the target samples. 
Taking inspiration from the modern semi-supervised learning techniques, \cite{tarvainen2017mean,berthelot2019mixmatch} where loss from unsupervised samples is scaled up, we introduce a schedule for monotonically scaling coefficient $\lambda$ for $T$ epochs as $\lambda(t) = \lambda_s + (\lambda_f - \lambda_s).s(t)$,
where, $\lambda_s$ is the scale at the first epoch $t=1$ and $\lambda_f$ is the scale at the epoch $t=T$.  
With the increase in the importance of the target samples, the network pulls target samples close to the source centroids and aligns the domains. The Gaussian Mixture model gradually adapts from a source only classifier to a domain invariant classifier. 
A Semantic diagram of the proposed approach is depicted in Figure \ref{model} which shows the flow of the source and the target samples through the neural network. The samples from source domain are used to train the network in a supervised way. The target samples are passed through the network to obtain their pseudo labels. The obtained pseudo labels are refined through our series of filters to obtain the most precise pseudo labels. These pseudo labels are used for training the network on target domain. 
We refer to our approach as DAPL (Domain Adaptation Using Pseudo Labels) in the paper.

\section{Experiments}

\subsection{Datasets and Experiments Setup}
\noindent We follow the standard setting for unsupervised domain adaptation, where the source domain consists of source samples and their labels and the target domain consists of unlabeled samples only. 

\noindent \textbf{VisDA} is a large dataset based on simulating Synthetic-to-Real condition. We follow the standard protocol as \cite{saito2018maximum}.
For this dataset, we use the standard ResNet-101 pretrained on ImageNet as the feature extractor. 
We add one fully connected layer followed by a Gaussian Mixture loss layer (classification layer) at the end of this network. We use stochastic gradient descent (SGD) optimizer with a momentum of 0.9 and a constant learning rate of 1e-4 for the added layers and 1e-6 for the feature extractor. 
We use only horizontal flip as the data augmentation. 

\noindent \textbf{Office-Home}\cite{venkateswara2017deep} experiments follows standard procedure as \cite{long2018conditional}.
We use ResNet-50 pre-trained on ImageNet as the backbone feature extractor for this dataset. 
Similar to VisDA, we add one fully connected and the Gaussian Mixture loss layer. This network is trained using stochastic gradient descent (SGD) optimizer with a momentum of 0.9 and a learning rate of 1e-2 for the classification layers and 1e-4 for the feature extractor. 
Random crop and Horizontal flip data augmentations are used.

\noindent \textbf{Digits} We use three digits datasets: MNIST, USPS and SVHN. 
For digit-related the experiments, we use architecture and protocol from \cite{bhushan2018deepjdot} and replace the last classification layer with the Gaussian Mixture loss layer. 
All the images are scaled to $32\times 32$ and the network is trained from scratch on the source dataset. We do not use any data augmentations in this experiment. 
we used Adam optimizer with learning rate of $1 \times 10^{-3}$ and $\beta_1 = 0.8$ and $\beta_2 = 0.999$. 
For MNIST$\rightarrow$SVHN, we also used class balance loss from \cite{french2018self} task to take care of the class imbalance issue in SVHN. Class imbalance loss is common in UDA and has been used in other methods as well like \cite{french2018self,lee2019drop} etc to handle this problem. It encourages networks to align the mean prediction of the mini-batch with uniform distribution. 

We implement a max-margin Gaussian mixture with margin $m=1$ and set $\gamma=0.1$ as per \cite{wan2018rethinking}.
We set $\lambda_s$ and $\lambda_f$ to $30$ and $150$ respectively for all our experiments to have significant contribution to updating the network’s parameters. 
$\tau_s$ is adjusted such that we have a sufficient number of $\bar{Y}$ for all the classes. It ensures that there are pseudo labels present for all the classes after the \emph{Confidence} filter.

\begin{table}[t]
\begin{center}
\scriptsize
 \begin{tabular}{l|c|c|c|c} 
 \toprule
Method & M$\rightarrow$U & U$\rightarrow$M & S$\rightarrow$M & M$\rightarrow$S\\ 
 \midrule
 Source only & 94.8 & 59.6  & 60.7 &33.4   \\
MMD\cite{long2015learning} & 88.5 & 73.5 & 64.8& -  \\
DANN\cite{ganin2016domain} & 95.7 & 90.0 & 70.8 & - \\
SE\cite{french2018self}& 88.1 & 92.4 & 93.3 & 33.9 \\
CDAN + E \cite{long2018conditional} & 95.6 & 98.0 & 89.2  & -\\
ATT \cite{saito2017asymmetric}& - & - & 85.0 & 52.8\\
DeepJDOT\cite{bhushan2018deepjdot}& 95.7 & 96.4 & 96.7  \\
PFAN\cite{chen2019progressive} & 95.0 & - & 93.9 & 57.6 \\
DIRT-T\cite{shu2018dirt} & - & - & 99.4 & \textbf{76.5}\\
ADR\cite{saito2018adversarial}& 91.3 & 91.5 & 94.1 & - \\
TPN\cite{pan2019transferrable}& 92.1 & 94.1 & 93.0  & -\\
ALDA\cite{chen2020adversarial} & 95.6 & 98.6 & 98.7  & -\\
DMRL\cite{wu2020dual} & 96.1 &\textbf{99.0} &96.2 & -\\
DAPL(Ours) & \textbf{98.8} & 98.6 & \textbf{98.9} & 66.5\\
\bottomrule
\end{tabular}
\end{center}
\caption{Results on Digits dataset: MNIST(M)$\leftrightarrow$USPS(U) and SVHN(S)$\leftrightarrow$MNIST(M).}
\label{digitresults}
\end{table}

\begin{figure*}[t]
    \begin{center}
       \includegraphics[width = 0.85\textwidth]{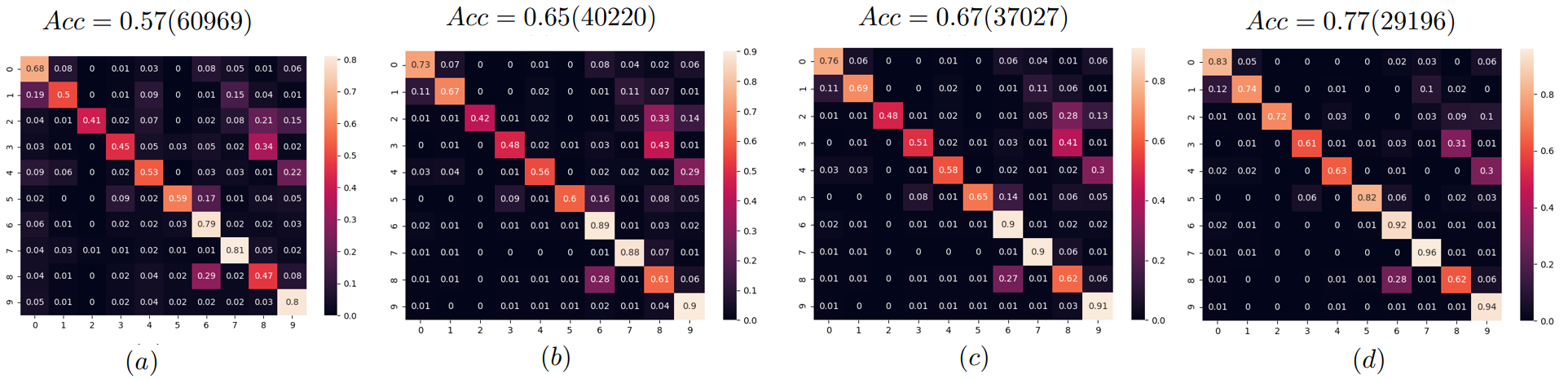}
    \end{center}
\caption{Confusion matrix after different filters for MNIST$\rightarrow$SVHN at $90\%$ training. (a) Unfiltered samples selected using Target Supervision Schedule, (b) after \emph{Confidence} filter, (c) after \emph{Conformity} filter, (d) after \emph{Consistency} filter. The numbers at the top indicate target classification accuracy with the number of target pseudo labels within parentheses.} 
\label{Fig:cm}
\end{figure*}

\subsection{Results}
\noindent In this section, we present results of our approach and compare them with the unsupervised domain adaptation techniques like ALDA \cite{chen2020adversarial}, DTA \cite{lee2019drop}, TADA \cite{wang2019transferable} etc. 
The results for the VisDA, Office-Home dataset and digits datasets are displayed in Table \ref{visdaresults}, Table  \ref{oh_results} and Table \ref{digitresults} respectively. 

\subsubsection{Digits} We tested our approach on different experiments for MNIST $\leftrightarrow$ USPS and SVHN $\leftrightarrow$ MNIST. Our method achieves high performance for all the combinations. 
It yields an accuracy of around $98\%$ for easy tasks - MNIST$\leftrightarrow$USPS and SVHN$\rightarrow$MNIST. MNIST$\rightarrow$SVHN is a difficult experiment as the domains are very far apart. 
DIRT-T \cite{shu2018dirt} outperforms our approach for this combination by using an instance normalization layer to process the input. Without it, the accuracy drops to $54.5\%$ which is lower than our method.

\subsubsection{VisDA \& Office-Home} Among the compared techniques, DAPL showcases a high performance for both the datasets. 
DAPL outperforms not only all the self-training techniques (ALDA \cite{chen2020adversarial}, DTA \cite{lee2019drop}, CDAN \cite{long2018conditional}, ADR \cite{saito2018adversarial}) which are similar to it, but also adversarial feature alignment approaches (DANN \cite{ganin2016domain},  MCD\cite{saito2018maximum}) and distance-metric based techniques (MMD \cite{long2015learning},  SWD\cite{lee2019sliced}). 
This showcases that pseudo label training can perform well when aided with accurate pseudo labels.

\section{Analysis}
\subsection{Ablation Analysis}
\noindent Here, we perform the ablation analysis of our approach and showcase how various components in our method enhance the performance of the network on the target samples. 
We use the SVHN$\leftrightarrow$MNIST case to demonstrate how different components behave under various settings. 
The results of this study can be viewed in Table \ref{ablationtable}.
We found Gaussian Mixture and Cross-Entropy to achieve comparabale performance.
We can observe all filters play a vital role in final performance of DAPL and removing any filter degrades the performance. 
Without a filter, there is an increase in the number of selected samples but the accuracy of the pseudo labels is adversely affected.

\begin{table}[t]
\begin{center}
\footnotesize
 \begin{tabular}{c|c|c} 
 \toprule
 Method & SVHN$\rightarrow$MNIST & MNIST$\rightarrow$SVHN \\ 
\midrule
 Cross-Entropy loss & 60.7  & 32.3 \\
 Gaussian Mixture loss & 61.5 & 31.5 \\
\midrule
\emph{Confidence} Only & 95.7 & 45.7 \\
\emph{Confidence} \& \emph{Conformity} & 96.9 & 49.7 \\
\emph{Confidence} \& \emph{Consistency} & 97.6 & 56.3 \\
\midrule
 All 3 filters & \textbf{98.9} & \textbf{66.5}\\
 \bottomrule
\end{tabular}
\end{center}
\caption{Results of the Ablation Analysis of the DAPL for MNIST$\leftrightarrow$SVHN.}
\label{ablationtable}
\end{table}

\subsection{Confusion Matrix}
\noindent Figure \ref{Fig:cm} depicts the confusion matrix for target classification for MNIST$\rightarrow$SVHN when $90\%$ of the training has been completed. 
It can be noted that there is an increase in accuracy after every filter. 
Our filters are able to increase the accuracy from $57\%$ to $77\%$ after discarding half the samples, confirming the capability of our filtration technique. 
We also show how the pseudo-label accuracy and the number of filtered samples vary over time in Figure \ref{Fig:cm} for Real-World$\rightarrow$Clipart experiment of the Office-Home dataset.

\begin{figure}[ht]
    \begin{center}
      \includegraphics[width=0.35\textwidth]{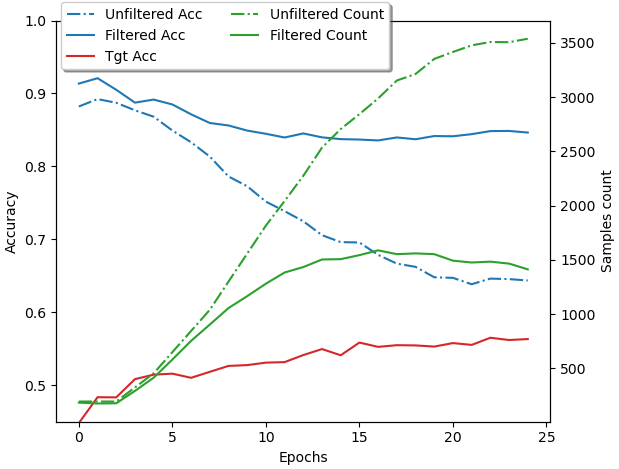} 
    \end{center}
\caption{Training plots for Real-World $\rightarrow$ Clipart experiment from Office-Home dataset. Unfiltered-Acc and Filtered-Acc are accuracies of pseudo labels without filtering (after Target Supervision Schedule) and with DAPL filtering, respectively. Tgt-Acc is the achieved target accuracy with DAPL. Unfiltered-Count and Filtered-Count are the number of selected pseudo labels without filtering and with DAPL filtering, respectively.}
\label{Fig:acc}
\end{figure}

\subsection{Pseudo Label Accuracy}
\noindent Figure \ref{Fig:acc} depicts the training plots for Real-World $\rightarrow$ Clipart experiment of Office-Home dataset.
In the absence of any filtering (Unfiltered Acc) the veracity of the pseudo labels (measured through accuracy) drops with training demonstrating confirmation bias. 
With DAPL filtering (Filtered Acc), the veracity remains steady.   
DAPL trades off quantity for quality as it selects fewer samples (Filtered Count) compared to the Unfiltered Count resulting in a steady climb in target accuracy (Tgt Acc).


\subsection{DAPL vs. Domain Alignment Loss}
\noindent Popular unsupervised domain adaptation techniques deploy a domain alignment loss (DAL) to align the domains. 
Instead, we rely on pseudo-label training for network generalization. 
Feature-based domain alignment techniques align the marginal distributions of the domains while being category-agnostic. 
This could lead to misalignment of the domains which in turn leads to poor target accuracies.

\begin{table}[t]
\footnotesize
\begin{center}
 \begin{tabular}{l|c|c} 
 \toprule
& \multicolumn{2}{c}{Initalization} \\ 
\midrule
Method & Source-Only & DANN \\
 \midrule
Initial & 34.6 & 50.9 \\
\midrule
DAPL + DAL & 78.4 & 75.9\\
\midrule
DAPL only & \textbf{ 81.9} & 79.6\\
\bottomrule
\end{tabular}
\end{center}
\caption{Comparative analysis of our approach(DAPL) with and without domain alignment loss (DAL) using VisDA dataset.}
\label{source_analysis}
\end{table}

\begin{figure}
    \begin{center}
      \includegraphics[width=0.38\textwidth]{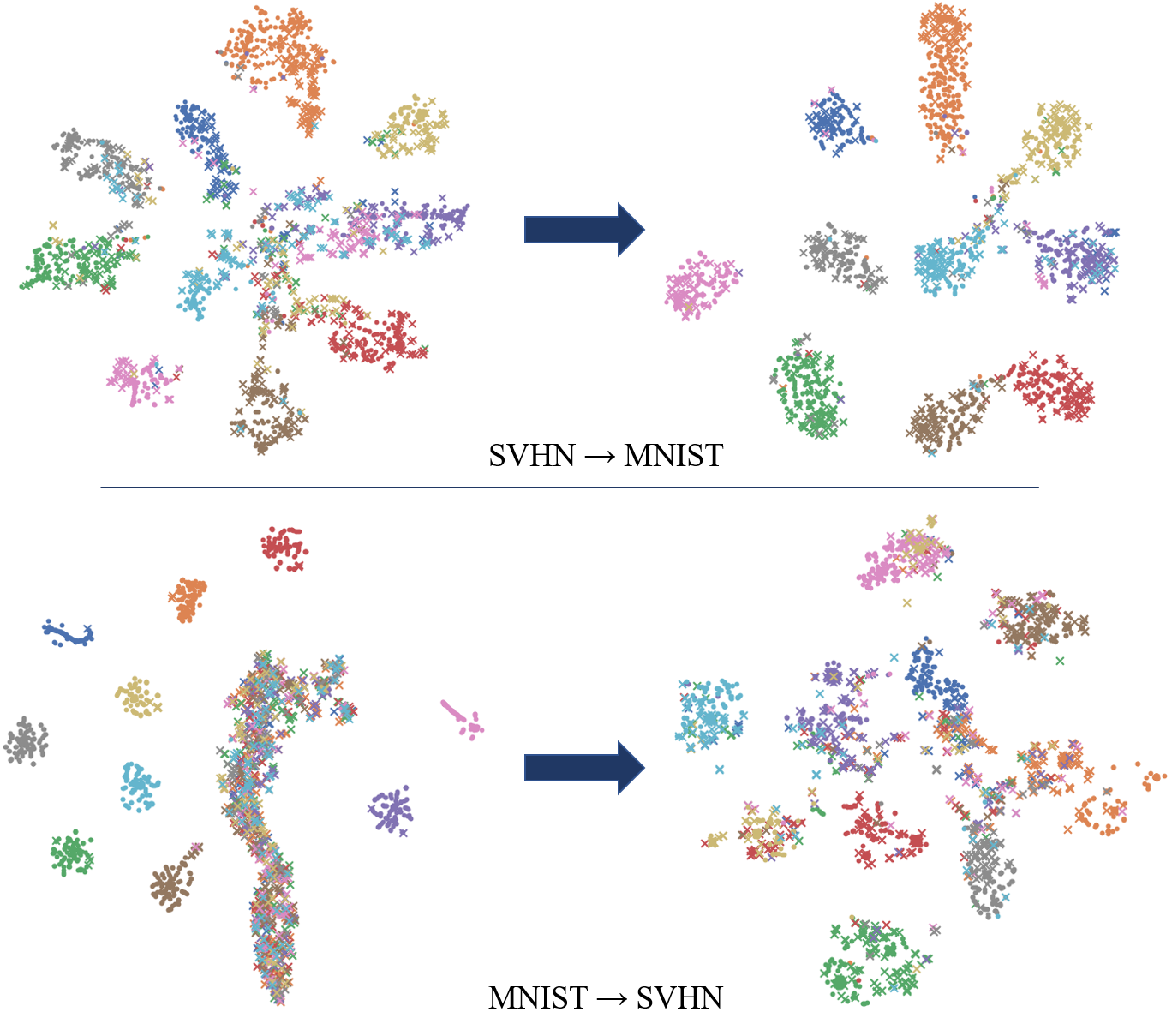}
    \end{center}
\caption{Visualization of image features using t-SNE. The source domain is represented using $\bullet$ and the target domain using $\mathbf{\times}$. Different colors depict different categories. Left column represents the source only features and the right column displays features after our approach. Best viewed in color.}
\label{digits_tsne}
\end{figure}

In Table \ref{source_analysis}, using the VisDA dataset, we compare DAPL with DANN \cite{ganin2016domain}, a popular adversarial feature-based domain alignment method. 
Column 1 (Source-Only) represents target accuracies for a network whose parameters have been initialized by training it with only source data. 
Column 2 (DANN) represents target accuracies for a network whose parameters have been initialized by training it on source data and a DANN loss to align the source and target. 
The `Initial' row depicts that DANN outperforms Source-only. 

The ``Initial'' two networks are further re-trained using: DAPL and DAPL + DAL, where DAL is marginal feature alignment (just like in DANN). 
Although the DANN initialization provides a higher accuracy at the start, it still results in a lower final performance in all the cases. 
This is because the classes are already mixed up by the DANN initialization. 
For the same reason, DAPL alone outperforms training the network with DAPL + DAL. 
The highest performance is achieved when no domain alignment loss is used, demonstrating how category agnostic domain alignment can degrade the network's performance on the target domain.

\subsection{Feature Visualization}
\noindent We use t-SNE plots to visualize the deep features of the VisDA dataset learned by Source-Only, DANN, and DAPL. 
From Figure \ref{Fig:tsne}, it is clear that the target data (colored) is scattered in the center initially and DAPL is able to cluster it into different classes.
Our approach creates compact clusters of the classes leading to alignment without any explicit use of a domain alignment loss.
We also showcase the deep features for MNIST$\leftrightarrow$SVHN task in Figure \ref{digits_tsne}.

\section{Conclusions}
In this paper we proposed a multi-stage pseudo-label filtering procedure for domain adaptation; a throwback approach relying upon the power of a pretrained network to extract generic features. 
Our approach selects a subset of most appropriate pseudo labels and using these pseudo labels, it gradually adapt the source classifier to accurately classify the target domain as well. 
Our experimental analysis evaluated different aspects of our approach and conclusively demonstrated the efficacy of our procedure against complex domain alignment approaches. Our approach outperforms all the compared approaches despite its simplicity on various datasets. 
However, we found that DAPL can underperform when the target samples size is small. In that case, our approach results in selecting a very small subset and does not adapt to target domain effectively. 
In the future work, we propose to refine the incorrect samples filtered out by DAPL. This way, we can include more pseudo labels for the target training.
{
    \small
    \bibliographystyle{ieeenat_fullname}
    \bibliography{main}
}


\end{document}